\documentclass{article}
\usepackage{ijcai23}

\usepackage{times}
\usepackage{soul}
\usepackage{url}
\usepackage[hidelinks]{hyperref}
\usepackage[utf8]{inputenc}
\usepackage[small]{caption}
\usepackage{graphicx}
\usepackage{amsmath}
\usepackage{amsthm}
\usepackage{booktabs}
\usepackage{algorithm}
\usepackage{algorithmic}
\usepackage[switch]{lineno}
\newcommand{\citet}[1] {\citeauthor{#1}~\shortcite{#1}}

\usepackage{latexsym}
\usepackage{cleveref}

\usepackage{amsmath,amsfonts,bm}

\def\eqref#1{equation~\ref{#1}}

\def\1{\bm{1}}

\def\mW{{\bm{W}}}

\DeclareMathAlphabet{\mathsfit}{\encodingdefault}{\sfdefault}{m}{sl}
\SetMathAlphabet{\mathsfit}{bold}{\encodingdefault}{\sfdefault}{bx}{n}

\usepackage[T1]{fontenc}    %
\usepackage{amsfonts}       %
\usepackage{nicefrac}       %
\usepackage{microtype}      %
\usepackage{enumerate}
\usepackage{graphicx}
\usepackage[export]{adjustbox}
\usepackage{booktabs}
\usepackage{tabu}
\usepackage{tikz}
\usepackage{tikzsymbols}
\usetikzlibrary{arrows.meta}
\usetikzlibrary{shapes.misc, positioning}
\usepackage{pgfplots}
\usepgfplotslibrary{groupplots}
\usepackage{multirow}
\usepackage{wrapfig}
\usepackage{subcaption}
\usepackage[toc,page]{appendix}
\usepackage[normalem]{ulem}
\usepackage{paralist}
\usepackage{cleveref}
\usepackage{makecell}
\usepackage{enumitem}
\usepackage{xspace}
\usepackage{array}
\usepackage{bbm}
\usepackage{enumitem}
\usepackage{pifont}
\usepackage{braket}

\newcommand{\ours}{{\textsc{Cp-NER}}}
\definecolor{myblue}{HTML}{6C8EBF}
\definecolor{mygreen}{HTML}{82B366}
\definecolor{myred}{HTML}{b85450}

\newtheorem{remark}{\noindent \textbf{Remark}}

\def\mW{{\bm{W}}}

\newcommand{\RN}[1]{%
	\textup{\lowercase\expandafter{\it \romannumeral#1}}%
}

\title{One Model for All Domains: \\ Collaborative Domain-Prefix Tuning for Cross-Domain NER}

\author{
Xiang Chen$^1$ \and
Lei Li$^1$ \and
Shuofei Qiao$^1$ \and
Ningyu Zhang$^1$*
\and
Chuanqi Tan$^3$\and \\
Yong Jiang$^3$ \and
Fei Huang$^3$ \And 
Huajun Chen$^{1,2}$*
\\
\affiliations
$^1$Zhejiang University \& AZFT Joint Lab for Knowledge Engine \\
$^2$Donghai Laboratory \\
$^3$DAMO Academy, Alibaba Group \\
\emails
\{xiang\_chen, leili21, shuofei, zhangningyu, huajunsir\}@zju.edu.cn\\
\{chuanqi.tcq, jiangyong.ml, f.huang\}@alibaba-inc.com
}

\begin{document}

\maketitle

\begin{abstract}

Cross-domain NER is a challenging task to address the low-resource problem in practical scenarios. Previous typical solutions mainly obtain a NER model by pre-trained language models (PLMs) with data from a rich-resource domain and adapt it to the target domain. Owing to the mismatch issue among entity types in different domains, previous approaches normally tune all parameters of PLMs, ending up with an entirely new NER model for each domain. Moreover, current models only focus on leveraging knowledge in one general source domain while failing to successfully transfer knowledge from multiple sources to the target. To address these issues, we introduce \textbf{C}ollaborative Domain-\textbf{P}refix Tuning for cross-domain \textbf{NER} (\textbf{\ours}) based on text-to-text generative PLMs. Specifically, we present text-to-text generation grounding domain-related instructors to transfer knowledge to new domain NER tasks without structural modifications. We utilize frozen PLMs and conduct collaborative domain-prefix tuning to stimulate the potential of PLMs to handle NER tasks across various domains. Experimental results on the Cross-NER benchmark show that the proposed approach has flexible transfer ability and performs better on both one-source and multiple-source cross-domain NER tasks.

\end{abstract}

\section{Introduction}

Named entity recognition (NER) is an important task for Knowledge Graph (KG) construction \cite{zhang2022deepke} and natural language processing (NLP).
Owing to the data scarcity issue in practical scenarios, obtaining adequate domain-related data is usually labour-intensive.
The naive idea of training models with rich-resource domain data (source) and transferring knowledge to a new specific domain (target) may struggle handing the semantic gap and limited data problem.
Hence, cross-domain NER, capable of learning information from the source domain to specific target domains with limited data, has been proposed to alleviate this issue.

\begin{figure}[!tpb] 
  \centering
    \includegraphics[width=0.4\textwidth]{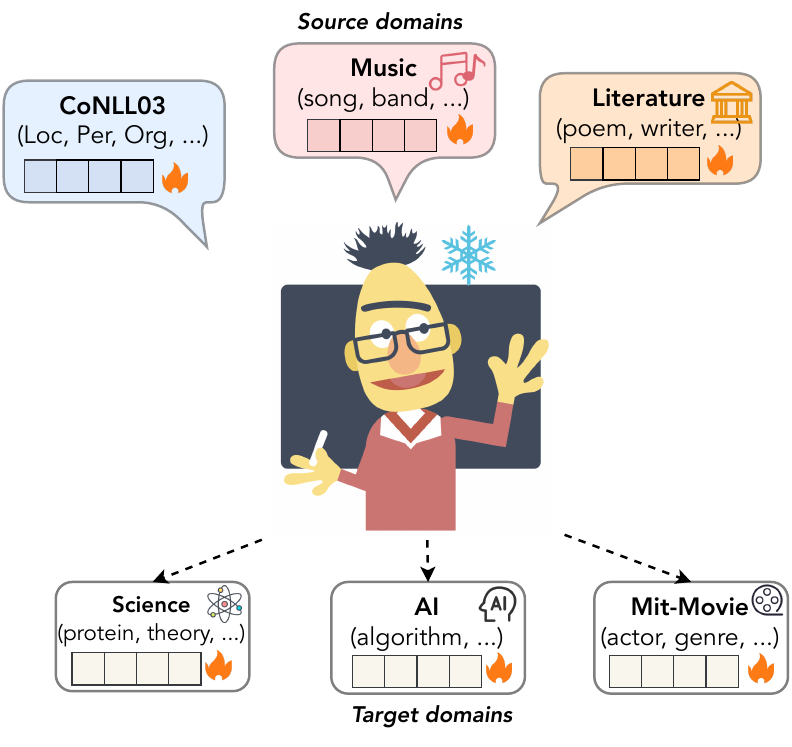}
    \caption{
        One  model for all domains with collaborative prefixes. 
    }
    \label{fig:motivation}
\end{figure}%

To address this task, some previous studies utilize adding auxiliary objects~\cite{liuzeroresource,wang2020-multi-domain} or designing new model architectures~\cite{jia2019cross,liu20-coach,jia2020multi} to train with both source and target domain data.
\citet{NER-BERT} leverages continue pre-training on the target domain for a better understanding of the domain-specific data.
Another line of researches~\cite{zheng2022-LSTNER,hu2022-LANER} focus on modeling the label relationship across domains to improve label information transfer.
Typically, \textsc{LANER}~\cite{hu2022-LANER}, the current state-of-the-art (SOTA) method on CrossNER benchmark~\cite{liu2021crossner}, utilizes an architecture to better leverage the semantic relationships among domains for boosting the cross-domain performance.

Despite the empirical success of previous works, several issues remain which have not been be appropriately solved.
\textbf{Firstly}, previous methods often rely on task-specific architectures for various domains with different entity categories.
For instance, \textsc{LANER}~\cite{hu2022-LANER} designs different model architectures to adapt a PLM to other target domains, which restricts the model's usability in more applications.
\textbf{Secondly}, most current methods are computationally inefficient and require tuning all parameters of the PLMs, which ends up with an entirely new NER model for each domain. 
For example, storing a full copy of pretrained
BERT\_Large~\cite{devlin2019bert} (340M parameters) for each domain is
non-trivial, not to mention the billion-scale large LMs. 
We introduce a summary of the above two points \textbf{from the perspective of practical applications}:
\textit{when a new domain data comes, it is not practical for users to design model architectures and tune all parameters in terms of computational resources and time.}
\textbf{Thirdly}, semantic knowledge transfer is essential for cross-domain NER, but previous works typically can only transfer from single source domain to the target domain and lack the ability of utilizing knowledge from multiple domains to help each other.

All the considerations above lay down our goal to investigate simple yet effective ways to keep one frozen set of PLM parameters for all domains and synthesize knowledge from multiple domains to enhance the target domain performance, as shown in Figure~\ref{fig:motivation}.
In this work, we enable collaborative domain-prefix tuning for cross-domain NER tasks.
As an alternative to fine-tuning the whole PLM, our method, namely {\ours}\footnote{Codes will be available in \url{https://github.com/zjunlp/DeepKE/tree/main/example/ner/cross}.}, tunes no existing PLM parameters. 
We establish this by casting the NER into a ``text-to-text generation grounding domain-related instructor''. 
Compared with \textsc{LightNER}~\cite{lightner} that generates indexes of entity spans, our formulation enables eliciting general knowledge about named entities from PLMs and avoids the external modification of model architecture. 
By introducing collaborative domain-prefix tuning consisting of a few trainable parameters, we can flexibly  adapt the knowledge from multiple source domains to target domains.
The specific process consists of three steps:
$(\RN{1})$ {\it Domain-specific Warmup.} 
This step leverages domain corpus to warm up the corresponding prefix; $(\RN{2})$ {\it Dual-query Domain Selector.}  
Identify which domain could benefit each other from both the perspectives of label similarity and prefix similarity;
$(\RN{3})$ {\it Intrinsic Decomposition for Collaborative Domain-Prefix.}
This module helps synthesize the final powerful prefix from multiple source domains.
In a nutshell, the contributions of our {\ours} are as follows:
\begin{itemize}

    \item We reformulate NER as the ``text-to-text generation grounding domain-related instructor''. 
    This new formulation helps trigger  PLMs to produce general knowledge about NER tasks and can handle different entity categories without any domain-specific modification of parameters for different domains, laying the groundwork of \textbf{one model for all domains}.

    \item Our proposal of collaborative domain-prefix tuning for PLMs aims to adapt the knowledge to cross-domain NER tasks.
    To the best of our knowledge, our work takes the first step to leverage knowledge transfer from multiple source domains specifically for cross-domain NER. 
    The comprehensive nature of the results is crucial for a wide range of cross-domain information extraction applications.

    \item We obtain SOTA  performance on the CrossNER benchmark involving transferring knowledge from a single general domain to the target domain. 
    Extensive experiments illustrate that our {\ours} can also efficiently leverage the knowledge transfer via learning lightweight parameters and unified architecture for all domains. 
\end{itemize}


\section{Related Work}

\subsection{Cross-Domain NER}
Although PLMs~\cite{devlin2019bert,liu2019roberta,lewis2020bart,xue2020mt5} have made waved in NLP, achieving overwhelming performance in widespread NER datasets, yet they require many domain-related labeled data for training when adapting to the target domains.
To this end, cross-domain NER algorithms~\cite{kim2015new,yang2018design,lee2018transfer,lin2018neural,liuzeroresource} that alleviate the data scarcity issue and boost the models' generalization ability to target domains have drawn substantial attention recently.
One approach to this problem is multitask learning, which includes adding auxiliary objects~\cite{liuzeroresource,wang2020-multi-domain} or designing new model architectures~\cite{jia2019cross,liu20-coach,jia2020multi} for improving the NER performance of the target domain by training on data from both source and target domain. 
However, these methods typically require training with vast amounts of labeled source domain data to obtain satisfactory target domain performance. 
Yet our approach leverages collaborative domain-prefix tuning, which allows for synthesizing parametric knowledge across domains and holds one frozen PLM for all domains.

Another area of research in cross-domain NER focuses on transferring label information across domains.
\citet{zheng2022-LSTNER} models the relation of labels as a probability distribution to better transfer cross-domain knowledge in NER.
\citet{hu2022-LANER} proposes an approach to better utilize the semantic relations among domains that jointly predict NER labels by the previous labels (from source) and corresponding token.
In contrast, our approach uses descriptive instructions with entity categories' semantics to enable the PLMs to understand NER tasks across different domains without modifying the model architecture. 
This allows for a single model to be used for all domains, laying the foundation for an efficient and versatile NER system.

\subsection{Parameter-Efficient Tuning for PLMs}

Pre-trained language models (PLMs)~\cite{devlin2019bert,liu2019roberta,lewis2020bart,xue2020mt5} have been widely used to achieve excellent performance in various NLP tasks~\cite{wang_glue:_2018,wang_superglue:_2019}. 
However, fine-tuning these models requires lots of computational resources and storage. 
To overcome this challenge, researchers have developed parameter-efficient methods to transfer PLMs to new tasks with small trainable parameters \cite{DBLP:journals/corr/abs-2203-06904}. 
These methods include prompt-tuning~\cite{shin-etal-2020-autoprompt,knowprompt}, which fine-tunes the model using task-specific prompts; and adaptors~\cite{pmlr-v97-houlsby19a,k-adapter,newman_p-adapters:_2021} which add task-specific parameters to the PLMs. 
However, applying these methods to new domains with unseen entity types can be challenging.
One type of solution is to utilize prefix-tuning \cite{DBLP:conf/acl/LiL20}, which prepends a sequence of task-specific vectors to the input.  
More works explore task adaptation with soft prompt~\cite{vu-etal-2022-spot,su-etal-2022-transferability}.

Different from those approaches, our work, referred to as {\ours}, stands out from those approaches in several ways. 
Firstly, we use collaborative knowledge updating between source and target prefixes to capture domain-related knowledge rather than just tuning extra parameters.
Secondly, our method allows for easy redeployment when switching to new domains with unseen entity types by using entity categories as grounding for text-to-text PLMs, unlike previous parameter-efficient methods. 
Finally, our approach aims to utilize multiple source domains to enhance the target domain, specifically focusing on NER, as opposed to transfer across tasks.
\section{Methodology}

\begin{figure*}[htpb] 
  \centering
    \includegraphics[width=1.0\textwidth]{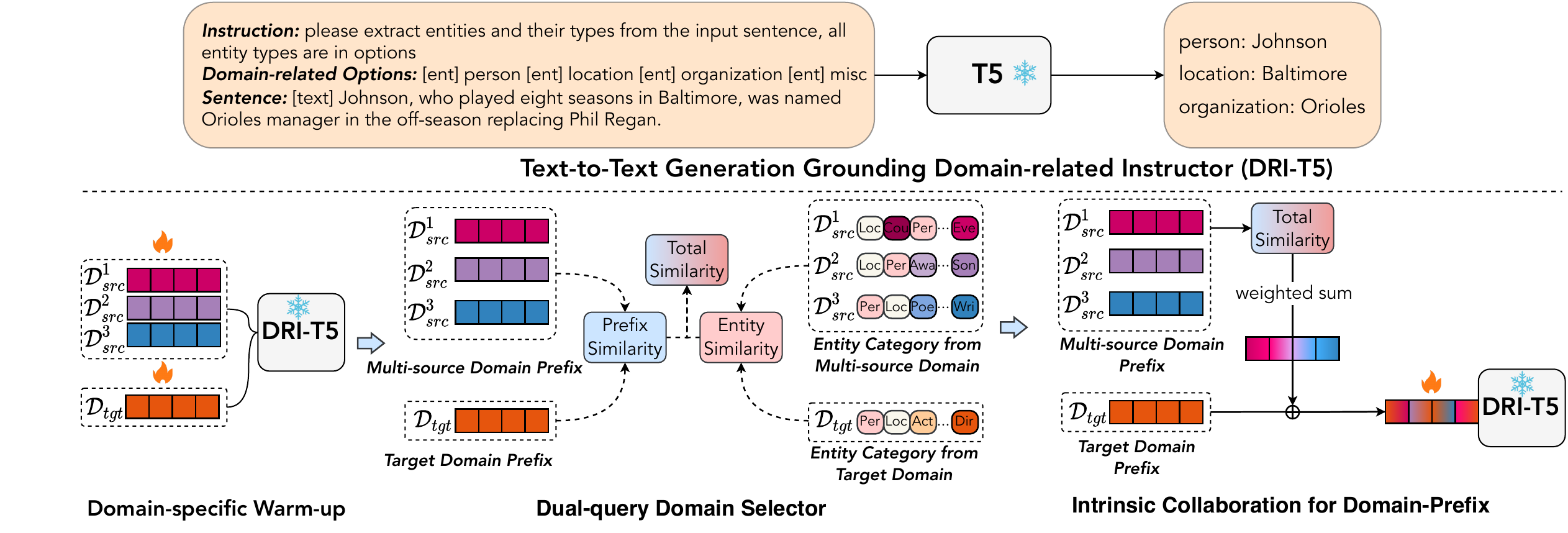}
    \caption{
        The overall architecture of the proposed {\ours} for cross-domain NER. 
        DRI denote the Domain-Related Instructor.}
    \label{fig:overview}
\end{figure*}%

\subsection{Problem Definition} 
\label{sec:definition}
Given sentence $\boldsymbol{x} = \{w_1, w_2, \dots, w_n\}$, the NER task is aimed to extracting all entities $\boldsymbol{y}$, and $n$ refers to the length of $\boldsymbol{x}$. 
Formally, we have the $i$-th entity $y_i \in \boldsymbol{y}$ as $y_i = (\boldsymbol{x}_{l:r}, t)$, where $t \in \mathcal{T}$ refers to the type 
 of entity.
$r$ and $l$ refers to the right and left entity boundary indexes, respectively.
Given the source domain data $\mathcal{D}_{src}$ as well as target domain data $\mathcal{D}_{tgt}$, the cross-domain NER task aims to obtain transferable knowledge with $\mathcal{D}_{src}$ to boosts the performance of $\mathcal{D}_{tgt}$. 
Previous studies on cross-domain NER mainly focus on transferring from one source domain to one target domain. While we not only consider ``\textbf{One Source for Target}'', but also include the ``\textbf{Multiple Source for Target}'' setting, which is practical for on-the-fly cross-domain NER in real scenarios.

\subsection{Text-to-Text  Generation Grounding Domain-related  Instructor} 
\label{sec:generation}
We formulate the cross-domain NER task as sequence-to-sequence generation and leverage a \textbf{frozen} T5 \cite{Raffel2020-T5}, as shown in Figure \ref{fig:overview}.
Specifically, for the NER task, we construct the inputs containing the following components:
\begin{itemize}
    \item \textbf{Sentence} - the input sentence $\boldsymbol{x}$.
    \item \textbf{Instruction} - the instruction $\boldsymbol(s)$ tells the model how to solve NER task. 
    We ask the model to generate a sequence according to the introduction. 
    \item \textbf{Domain-related Options} - the option ($\boldsymbol{o}$) includes all entity types $\mathcal{T}$ of the current domain and acts as both a constraint and a hint.
\end{itemize}
Formally,  {\ours} takes the given  instructor ($\boldsymbol{s}$), domain-related option ($o$) and the text sequence ($\boldsymbol{x}$) as input and generates the output ($\boldsymbol{y}$), which contains the extracted information from input sentence based on domain-related instructor:
\begin{equation} \label{equ:uie}
    \boldsymbol{y} = \mathrm{LM}_\theta(\boldsymbol{s} ; \boldsymbol{o} ; \boldsymbol{x}),
\end{equation}

where $\theta$ denotes parameters of the  ${\rm LM}_\theta$ (we adopt T5 model in this paper).
We convert the output sequence into a natural language consistent with the input instructions.
To be specific, we leverage simple templates to reformulate the entity $(\boldsymbol{x}_{l:r}, t)$ to natural language by concatenating entity types, mentions and several special tokens (separators).
For example, given the sentence ``Johnson, who played eight seasons in Baltimore, was named Orioles manager in the off-season replacing Phil Regan.'', the model will generate ``((person: Johnson) (location: Baltimore) (organization: Oriloes))''.

We argue that the text-to-text generation grounding domain-related instructor can:
1) effectively guide the T5 model to generate named entity sequence with \textbf{domain-specific entity categories} so that it can be transferred to new domain NER tasks without any structural modification of the T5 model;
2) stimulate the potential of PLM to handle NER tasks with various domains, which laid the foundation for guiding frozen PLM to generate entity sequences through prefix-tuning.

\subsection{Insights of Tuned Prefix as Domain Controller}
\label{sec:prefix}
Prefix-tuning~\cite{li2021prefix} prepends trainable continuous tokens, also known as \textit{soft prompts}, to each Transformer layer's hidden states.
We regard the prefix as the domain controller for cross-domain NER.

As described in ~\citet{li2021prefix}, directly updating the prefix parameters leads to unstable optimization and a slight drop in performance.
So the prefix of the $j$-th layer, $\mathbf{\delta}^{(j)}$, is obtained based on the trainable matrix, $\mathbf{P}^{(j)}$, while all the other parameters of PLM will be fixed. 
In particular, we derive the variant formula of self-attention with the prefix at the $j$-th layer as\footnote{Without loss of generalization, we ignore the scaling factor $\sqrt{d}$ of the Softmax for the ease of representation.}:

\begin{equation}
\label{eq:prefix-adapter}
\small
\begin{split}
& \mathrm{head}^{(j)}  = \text{softmax}\big( \textbf{x}^{(j)}\mW_q^{(j)} [\delta_{k}^{(j)}; \textbf{x}^{(j)}\mW_k^{(j)}] ^\top\big) \begin{bmatrix}  \delta_{v}^{(j)} \\ \textbf{x}^{(j)}\mW_v^{(j)} \end{bmatrix} \\
& = (1 - \lambda(\textbf{x}^{(j)})) \underbrace{ \text{Attn}(\textbf{x}^{(j)}\mW_q^{(j)}, \textbf{x}^{(j)}\mW_k^{(j)}, \textbf{x}^{(j)}\mW_v^{(j)}) }_{\text{standard attention}} \\
&+ \lambda(\textbf{x}^{(j)}) \underbrace{ \text{Attn}(\textbf{x}^{(j)}\mW_q^{(j)}, \delta_{k}^{(j)},\delta_{v}^{(j)}) }_{\text{attention of domain-specific prefix }},
\end{split}
\end{equation}
where $\lambda(\textbf{x}^{(j)})$ refers to the scalar regarding the sum of normalized attention weights on the key and value embedding from prefixes.

For simplicity, we denote the prefix indexes as $\mathrm{P}_{\rm idx}$, then prefix-tuning~\cite{li2021prefix} formulates the forward propagation at the position $n$ as:
\begin{equation}
    h_{n}^{(j+1)} = \mathrm{LM}_\theta^{(j)}\left(h_{n}^{(j)} ~\middle|~ h_{<n}^{(j)} \right),
\label{prefix-propagate}
\end{equation}
where $h_i$ is an activation of the trainable parameter $\mathbf{P}$, $z = [\text{PREFIX}, \boldsymbol{x}$],
$h_{i}^{(j)} = {\delta}^{(j)}[i, :]$ for all $j=0$ to $L-1$, $i \in \mathrm{P}_{\rm idx}$ and $h_{i}^{(0)} = z_{i}$ for $i \notin \mathrm{P}_{\rm idx}$.
With the large-scale LM parameters fixed and continuous optimization on training samples, prefix-tuning is expected to steer the LMs to generate correct entity prediction for test data.

\begin{remark}
As shown in Eq.~\ref{eq:prefix-adapter}, the domain-specific prefix tuning essentially modifies the original head attention through linear interpolation by a scalar factor (i.e., $1-\lambda$), acting as a domain controller to drive the model to output domain-specific predictions. However, prefix-tuning still suffers from knowledge forgotten or dilution in the process of direct domain transfer, which inspires us to further explore more effective ways of joining prefix hints to leverage knowledge from multiple source domains.
\end{remark}

\subsection{Collaborative Domain-prefix Tuning}
\label{sec:colla-prefix}

Directly transferring knowledge through prefix-tuning can be challenging because the knowledge learned in the source prefix can easily be forgotten or diluted when fine-tuned on data from the target domain due to differences in text styles and entity categories. 
To address this issue and help the T5 model's domain controller more effectively capture knowledge from source domains, we propose a collaborative domain-prefix tuning method.

The overall process is illustrated in Figure~\ref{fig:overview}: (1) domain-related warm-up to capture knowledge from the corpus of the current domain; (2) dual-query domain selector to determine the proportion of prefix knowledge from multiple source domains; (3) intrinsic decomposition for collaborative domain-prefix, which can flexibly synthesize prefix knowledge from source and target domains.

\paragraph{Domain-specific warm-up.}
To flexibly organize prefixes with knowledge from different domains, we first leverage domain corpus to warm up the corresponding prefixes.
Formally, we warm up the newly initialized trainable parameter matrix $\{\mathbf{P}^{(0)}, \dots,\mathbf{P}^{(L-1)}\}$ for all $L$ layers of frozen T5 
where the final prefix of $l$-th layer is drawn from the parameter matrix $\mathbf{P}^{(l)}$.
We optimize the parameter matrix of prefix at each layer $\{\mathbf{P}^{(0)}, \dots, \mathbf{P}^{(L-1)}\}$ using the training set $\mathcal{D}_{tr}$.
We follow the log-likelihood objective in T5~\cite{Raffel2020-T5} and abbreviate the optimization objective  as follows:
\begin{equation}
\label{eq:prefix-optimzation}
    \begin{aligned}
        & \min_{\{\mathbf{P}^{(0)}, \dots, \mathbf{P}^{(L-1)}\}} \mathbb{E}_{(x, y)\sim\mathcal{D}_{tr}} \left[\sigma \left(h_{n}^{(L)}, y\right) \right] 
    \end{aligned}
\end{equation}
where $\sigma$ as the $\mathrm{softmax}$ scoring function that output $h_o^{(L)}$ to a probability vector over the vocabulary. 
 $h_{i}^{(j)}$ refers to the intermediate activation of the $j$-th layer at step $i$.
We utilize the above optimization objective over the domain corpus to warm up the domain-specific parameter matrix and then generate the domain controller $\{\delta^{(0)}, \dots, \delta^{(L-1)}\}$.

\paragraph{Dual-query domain selector.}

When the source and target domains share part of the same entity categories, it is intuitive to leverage those shared labels having similar semantic information for flexible adaptation to the target domain. 
Besides, domain-specific prefixes maintain different grammar styles and themes.
Based on the above consideration, we design a dual-query domain selector from both the perspective of label similarity and prefix similarity to identify which domains could benefit each other. 

\begin{itemize}
\item[$\bullet$] 
\textbf{Entity Similarity:} we measure entity similarity through the embedding of the T5 model. 
Denote the word embedding of T5 as ${\rm LM}_w$; we feed entity categories of source and target domain into the ${\rm LM}_w$ to obtain the semantic category representations of the $i$-th source domain as $\mathrm{E}_{src}^{i} \in \mathbb{R}^{|\mathcal{T}^{i}_{src}| \times d}$ and target domain as $\mathrm{E}_{tgt} \in \mathbb{R}^{|\mathcal{T}_{tgt}| \times d}$.
Then we conduct pooling over the representation and calculate their similarity matrix as follows:
\begin{equation}
\mathcal{S}_e^i= \textit{ cos}(\textit{Pool}(\mathrm{E}_{src}^{i}),  \textit{Pool}(\mathrm{E}_{tgt})) ,
\end{equation}
where $\textit{Pool}$ is the pooling operation function and 
$\mathcal{S}_e^i \in \mathbb{R}$;

\item[$\bullet$] 
\textbf{Prefix Similarity:} given the prefix $\delta_{src}=\{\delta_{src}^{(0)}, \dots, \delta_{src}^{(L-1)}\}$ and $\delta_{tgt}=\{\delta_{tgt}^{(0)}, \dots, \delta_{tgt}^{(L-1)}\}$ of source and target domain, we also apply cosine similarity function over prefixes to calculate prefix similarity as:
\begin{equation}
\mathcal{S}_p^i= Avg( \textit{cos} ( \delta_{src}, \delta_{tgt}) ),
\end{equation}
where $\mathcal{S}_p^i \in \mathbb{R}$ \ and $Avg$ represents the average function.
\end{itemize}
For the target domain $\mathcal{D}_{tgt}$,  we denote $\alpha$ as the hyper-parameter $\alpha \in [0:1]$ to determine the ratio between the dual-query similarities as follows:
\begin{equation}
\mathcal{S}_{total}^i= \alpha^i \mathcal{S}_e^i + (1-\alpha^i) \mathcal{S}_p^i .
\end{equation}
As for leveraging $I$ source domains for collaborative prefix tuning, we apply the softmax function over the total similarity of multiple source domains as follows:
\begin{align}
    \mathbf{S}_{total} = \sigma ([\mathcal{S}_{total}^1;\dots;\mathcal{S}_{total}^I]),
\end{align}
To pursue compositional generalization and differentiable optimization, we encourage the prefix to be continuously activated and combined. 
Every single prefix from source domains can be viewed as some fundamental skill and domain-specific knowledge, and  NER tasks from another domain can be solved by combining such modular prefixes. 
Though similar ideas have been proposed in other names and contexts~\cite{Sun2020Sparse,Ponti2022Modular}, this is the first work that implements the cross-domain NER with multiple source domains to drive PLMs to recognize named entities.

\begin{table*}[htbp]
\renewcommand{\arraystretch}{1.0}
\centering
\small
\scalebox{0.82}{
\begin{tabular}{lcccccc}
\toprule
\multicolumn{1}{l}{\multirow{2}{*}{\textsc{\textbf{Models}}}} 
& \multicolumn{6}{|c}{\textsc{\textbf{Conll2003}}} \\
& \multicolumn{1}{|c}{\textsc{\textbf{Politics}}}  
& \textsc{\textbf{Science}}  
& \textsc{\textbf{Music}}     
& \textsc{\textbf{Literature}}
& \multicolumn{1}{c|}{\textsc{\textbf{AI}}}
& \textsc{\textbf{Average}}
\\ \midrule 
\cmidrule[0.12ex]{1-7}
\multicolumn{7}{l}{\textbf{w/o DAPT}} \\
\multicolumn{1}{l|}{\textsc{Coach}~\cite{liu20-coach}}       
& 61.50   & 52.09 & 51.66   &  48.35               & \multicolumn{1}{c|}{45.15}     & 51.75     \\
\multicolumn{1}{l|}{\textsc{Cross-Domain LM}~\cite{jia19-modeling}}    
& 68.44 & 64.31 & 63.56   & 59.59                  & \multicolumn{1}{c|}{53.70}     & 61.92     \\
\multicolumn{1}{l|}{\textsc{Flair}~\cite{Alan18-flair}}       
& 69.54  & 64.71 & 65.60   &  61.35                & \multicolumn{1}{c|}{52.48}     & 62.73     \\
\multicolumn{1}{l|}{\textsc{BARTNER-base}~\cite{BARTNER}}      
& 69.90   & 65.14 & 65.35  &  58.93                & \multicolumn{1}{c|}{53.00}     & 62.46     \\   

\multicolumn{1}{l|}{\textsc{LST-NER}~\cite{zheng2022-LSTNER}}      
& 70.44   & 66.83 & 72.08  &  67.12                & \multicolumn{1}{c|}{60.32}     & 67.36     \\  
\multicolumn{1}{l|}{\textsc{LANER}~\cite{hu2022-LANER}}     
& 71.65   &  69.29 & 73.07 &  67.98                & \multicolumn{1}{c|}{61.72}     & 68.74     \\
\multicolumn{1}{l|}{\textsc{LightNER}~\cite{lightner}}      
& 72.78   &  66.74 & 72.28 &  65.17                & \multicolumn{1}{c|}{35.82}     & 62.56     \\
\multicolumn{1}{l|}{\textbf{\ours}}
 & \textbf{73.41} & \textbf{74.65}       & \textbf{78.08} & \textbf{70.84} &\multicolumn{1}{c|}{\textbf{64.53}} & \textbf{72.30}   \\
 \cmidrule(){1-7}
 \multicolumn{7}{l}{\textbf{Introducing DAPT}} \\
 \multicolumn{1}{l|}{\textsc{DAPT}~\cite{liu2021crossner}} 
& 72.05 & 68.78 & 75.71 & 69.04      & \multicolumn{1}{c|}{{62.56}} & {69.63} \\  
\multicolumn{1}{l|}{\textsc{LST-NER}+DAPT}      
& 73.25   & 70.07 & 76.83  &  70.76                & \multicolumn{1}{c|}{63.28}     & 70.84     \\   
\multicolumn{1}{l|}{\textsc{LANER}+DAPT} 
& 74.06   &  71.83 &78.78  &  71.11                & \multicolumn{1}{c|}{65.79}     & 72.31    \\  
\multicolumn{1}{l|}{\textbf{\ours+DAPT}}
& \textbf{74.25} & \textbf{75.82} & \textbf{79.10} & \textbf{72.17}       & \multicolumn{1}{c|}{\textbf{67.95}} & \textbf{73.86} \\
\bottomrule
\end{tabular}
}
\caption{
Comparisons of existing studies and our CP-NER in terms of F1 scores are provided. 
The AVERAGE indicates the average F1 score across five domains in the CrossNER benchmark using only a single source (\textsc{{Conll2003}}). 
The baseline results are cited directly from the LANER and LST-NER papers. 
}
\label{tab:main_results}
\end{table*}

\paragraph{Intrinsic  collaboration for domain-prefix.}
Based on the selection of source domains to benefit target domains, we propose intrinsic collaboration for domain-prefix, which can synthesize the final powerful prefix from multiple source domains.
Specifically, for the warm-up parameter matrix $\mathbf{{P}}^{(j)}\in \mathbb{R}^d$ from the source domain, we argue it is in the same intrinsic space as the prefix from target domain.
We perform \textit{intrinsic refactor} such as addition operation
to the warm-up parameter matrices of the target domain, as shown in Figure~\ref{fig:overview}.  The specific parameter matrix $\mathbf{P}_{tgt}^{(j)}$ of target prefix at the $j$-th layer is refactored as: 
\begin{equation}
\hat{\mathbf{P}}_{src}^{(j)}=\mathcal{S}_{total} \cdot [\mathbf{P}_{src_1}^{(j)}; \dots; \mathbf{P}_{src_I}^{(j)}] ,
\end{equation}

\begin{equation}
\label{eq:lora}
    \hat{\mathbf{P}}_{tgt}^{(j)} \leftarrow \left(\mathbf{P}_{tgt}^{(j)} + \hat{\mathbf{P}}_{src}^{(j)}
    \right)/2,
\end{equation}
where $\hat{\mathbf{P}}_{src}^{(j)}$ denote the aggregated matrix from multiple domains\footnote{For scenarios with only one source domain, we let $\hat{\mathbf{P}}_{src}^{(j)}={\mathbf{P}}_{src}^{(j)}$ }. 
We further  follows the objective function in Eq.\ref{eq:prefix-optimzation} to update $\hat{\mathbf{P}}_{tgt}^{(j)}$  while keeping $\mathrm{LM}_\theta$ frozen.

\begin{remark}
From the optimal control (OC) perspective, prefix-tuning can be formalized as seeking the OC of the pre-trained LM for specific domains.
 And our collaborative domain-prefix tuning can be interpreted as seeking the close-loop control for leveraging knowledge from multiple-source to enhance the target domain NER performance.
\end{remark}

\section{Experiments}

\subsection{Experimental Settings}

\paragraph{Dataset.}

We conduct experiments using four publicly available datasets, including CrossNER~\cite{liu2021crossner}, CoNLL 2003~\cite{conll03}, MIT Restaurant~\cite{mitres-dataset}, and MIT Movie~\cite{mitrmovie-dataset}. 
CoNLL 2003 is a well-known NER dataset that serves as the source domain, which is labeled with four categories: PER, LOC, ORG, and MISC. 
Additionally, CrossNER contains five separate domain datasets: politics, natural science, music, AI, and literature. 
We adhere to the official splits for training, validation, and test sets, and the statistics for these datasets can be found in the supplementary materials.

\begin{table*}[t!]
\renewcommand{\arraystretch}{1.0}
\centering
\small
\scalebox{0.82}{
\begin{tabular}{c|l|c|c|ccc|ccc}
\toprule
{\multirow{2}{*}{\textsc{\textbf{Source}}}} 
& {\multirow{2}{*}{\textsc{\textbf{Methods}}}} 
& \multicolumn{1}{c|}{\textsc{\textbf{Sci.}}}
& \multicolumn{1}{c|}{\textsc{\textbf{AI}}}
& \multicolumn{3}{c|}{\textsc{\textbf{MIT Mov.}}}
& \multicolumn{3}{c}{\textsc{\textbf{MIT Res.}}}

\\
\cmidrule{3-10}
  & &All  & All &10 &20 &50  &10 &20 &50  \\

\midrule

    \multirow{4}{*}{\textsc{None}} 
    & \textsc{LANER}~\cite{hu2022-LANER}  
    & 67.79 & 58.65 
    & 49.25 & 54.48 & 70.96 & 45.15
    & 49.14 & 65.10 \\
    & \textsc{LightNER}~\cite{lightner} 
    & 49.94 & 15.02
    & 12.90 & 39.18 & 67.03 & 5.70
    & 23.03 & 51.15 \\
\cmidrule{2-10}
    & \textbf{\ours} 
    & \textbf{70.24}
    & \textbf{60.24}  
    & \textbf{60.01} 
    & \textbf{67.21} & \textbf{81.57}  & \textbf{49.42}
    & \textbf{63.24} & \textbf{75.92}  \\

\midrule
    \multirow{6}{*}{\textsc{Conll.}+
    \textsc{Pol.} +
    \textsc{Lit.} +
    \textsc{Mus.}} 
    & \textsc{LANER} <Chain Transfer>
    & 67.82 & 55.60 & 50.65 
    & 60.63 & 72.60 & 38.32 
    & 45.56 & 58.38  \\
    & \textsc{LANER} <Ensemble>
    & 69.97 & 59.47 & 53.55 
    & 63.01 & 72.85 & 44.79 
    & 50.99 & 62.30  \\
    & \textsc{LightNER} <Chain Transfer>
    & 67.49 & 40.91 & 51.77 
    & 66.19 & 74.44 & 48.13 
    & 52.97 & 65.13  \\
    & \textsc{LightNER} <Ensemble>
    & 58.69 & 26.25 & 32.01 
    & 52.63 & 67.29 & 26.50 
    & 57.09 & 67.07  \\
\cmidrule{2-10}
    & \textbf{\ours} <Multiple Source>
    & \textbf{75.33}  & \textbf{64.90} & \textbf{70.74} 
    & \textbf{73.22} & \textbf{81.90}  & \textbf{65.14}
    & \textbf{70.46} & \textbf{77.34}  \\

\bottomrule
  
\end{tabular}
}
\caption{\label{tab:cross}
Model performance, measured by the F1 score, in the setting of transferring from multiple source domains to the target domains.}
\end{table*}

\paragraph{Baselines.}

To evaluate the effectiveness of the proposed method, we compare it with several  baselines, including:

\begin{itemize}
\item[$\bullet$] \textsc{DAPT}~\cite{liu2021crossner}: DAPT employs a large unlabeled corpus related to a specific domain, which is based on BERT and fine-tunes it on the CrossNER benchmark.

\item[$\bullet$] \textsc{Coach}~\cite{liu20-coach}: This method utilizes patterns of slot entities and combines the features for each slot entity in order to improve the accuracy of entity type predictions.

\item[$\bullet$] \textsc{Cross-Domain LM}~\cite{jia19-modeling}: This method utilizes a parameter generation network to merge cross-domain language modeling with NER, resulting in improved model performance.

\item[$\bullet$] \textsc{Flair}~\cite{Alan18-flair}: This method utilizes the internal states of a character-level language model to generate contextual string embeddings, which are integrated into the NER model.

\item[$\bullet$] \textsc{BARTNER}~\cite{BARTNER}: This approach uses the pre-trained BART model to generate entity spans, treating the NER task as a sequence generation problem. 

\item[$\bullet$] \textsc{LST-NER}~\cite{zheng2022-LSTNER}: This approach models the relationship between labels as a probability distribution and builds label graphs in both the source and target label spaces for cross-domain NER tasks. 

\item[$\bullet$] \textsc{LANER}~\cite{hu2022-LANER}: This method introduces a new approach for cross-domain named entity recognition by utilizing an autoregressive framework to strengthen the connection between labels and tokens.

\item[$\bullet$] \textsc{LightNER}~\cite{lightner}: This utilizes a pluggable prompting method to improve NER performance in low-resource settings.

\end{itemize}
Due to all baselines being reported in \textbf{base PLMs}, we utilize \textbf{T5-base} to conduct a fair comparison for all experiments.

\subsection{Transfer from Single Source Domain}

As we can see from the \textbf{w/o DAPT} column of Table~\ref{tab:main_results}, {\ours} can consistently obtain better performance than SoTA method LANER across all target domains, with a 3.56\% F1-score improvement in the \textit{Source \& Target} setting.
Moreover, we find that {\ours} surpasses LANER on the science and music domains,  with a 6\% to 8\% F1-score performance gains on each domain. Experimental results demonstrate that our {\ours} technique achieves superior performance on the CrossNER benchmark. 
Our approach beats explicitly naive baselines and is on par with the LANER+DAPT model, which adopts an external domain-related corpus for pre-training.

On the other hand, we also apply DAPT on our {\ours} to conduct a fair comparison with other baselines. 
As \textbf{Introducing DAPT} column of Table~\ref{tab:main_results}  illustrates, our method ({\ours}) can better recognize and categorize entities on these domains with DAPT. 
This is because it leverages many domain-related data, which can boost performance.
When introducing DAPT, we observe that {\ours} also outperforms other baselines in all domains, revealing that our method is compatible with DAPT.

\begin{table*}[htp]
\small
\centering
 \scalebox{0.85}{
  \begin{tabular}{m{1.0cm}m{6cm}m{4cm}m{5cm}}
  \toprule
  Target & Sentence & Prediction & Reference Source \\
  \midrule
Mit-Movie & 
was vivien leigh in a \textbf{musical} that was rated r and received nine stars and above
& 
\makecell[l]{
None: <genre> Null  \\
Multi-Source: <genre> musical  \\
}
&
\textbf{Literature}: The following year they collaborated on a \textbf{musical} film version of The Little Prince , based on the classic children 's tale by Antoine de Saint-Exupry.
\\ \midrule
Science
& 
These include the TRUE Blue Crew, Antipodean Resistance, the Australian Defence League, National Action (\textbf{Australia}), the Q Society, Reclaim Australia and the Lads Society (formerly United Patriots Front).
 &
\makecell[l]{
None: <organization> Null  \\
Multi-Source: <organization> \\ Australiza  \\
}
&
\textbf{Music}: In addition to relentless touring in the U.S. and Canada, PUSA made multiple tours of Europe, \textbf{Australia}, New Zealand and Japan."
\\\midrule
AI 
& 
Other films between 2016 to 2020 that captured with IMAX camera's were Zack Snyder's Batman v Superman: Patty Jenkins' Wonder Woman 1984, Cary Joji Fukunaga's No Time to Die and \textbf{Joseph} Kosinski's Top Gun: Maverick.
 &
\makecell[l]{
None: <researcher> Joseph  \\
Multi-Source: <person> Joseph \\
}
&
\textbf{Literature}: Her stage credits include Norman Mailer's The Deer Park, Israel Horovitz's The Indian Wants the Bronx, Neil Simon's The Good Doctor and \textbf{Joseph} Papp's 1974 Richard III at the Lincoln Center.
\\
\bottomrule
 \end{tabular}
 }
\caption{Cases analysis between None source domain transfer and transfer with multiple source domains.}
\label{tab:case}
\end{table*}

\subsection{Transfer from Multiple Source Domains} 
Previous work for cross-domain NER only focuses on transferring from a single source domain to a target domain, failing to leverage multiple source domains, even though numerous source domains are more in line with practical application scenarios.
As shown in Table~\ref{tab:cross}, we experiment with transferring from multiple source domains. 
We use the CoNLL 2003, Politics, Literature, and Music datasets as source domains and test on the  Mit-Movie, AI, Science, and Mit-Restaurant target domains. To compare our {\ours} with other baselines, we propose the following setting for multiple source domains.

\paragraph{Setting.}
$(\RN{1})$ {\it <Ensemble>.}Train on a single source domain and transfer to the target domain, then ensemble the models transferred from multiple source domains.
$(\RN{2})$ {\it <Chain Transfer>.} Continuously train on multiple source domains in sequence and then transfer to the target domain.
$(\RN{3})$ {\it {\ours}<Multiple Source>.} Leverage the warm-up domain-specific prefix to conduct collaborative domain-prefix tuning.

\paragraph{Multiple source makes better transfer learning.}
According to the results of Table~\ref{tab:main_results} and Table~\ref{tab:cross}, it can be observed that using multiple sources provides consistent improvement for baseline models.
However, only \textsc{LightNER}<Ensemble> and \textsc{LANER}<Chain Transfer> achieve little improvements compared with the results in the single source domain setting, revealing that <Ensemble> may be more suitable for lightweight tuning. In contrast, <Chain Transfer> is ideal for fully-tuned models. 
Furthermore, our {\ours} using collaborative prefix tuning empowers the effectiveness of multiple sources and achieves excellent improvement compared to using a single source. 
The results in Table~\ref{tab:cross} show that our model outperforms all baseline methods in both the None source domain and multiple source domain settings, demonstrating that the proposed approach can be beneficial in transferring knowledge learned from the source domain.

\paragraph{Effect of data size.}

To investigate the influence of the size of the target domain data on performance, we conduct experiments using varying number of target training data (ranging from 10 to 50 samples) based on the settings that yielded the best results.
From Table ~\ref{tab:cross}, we notice that most of the models achieve performance improvement as the amount of data increases, indicating the essential role of labeled data.
Furthermore, it can be seen that our proposed method ({\ours}) consistently outperforms the baselines (\textsc{LANER} and \textsc{LightNER}) across all data sizes, further demonstrating the effectiveness of the approach in incorporating information from multiple source domains through collaborative domain-prefix tuning for cross-domain NER.

\begin{figure}[!tpb] 
  \centering
    \includegraphics[width=0.48\textwidth]{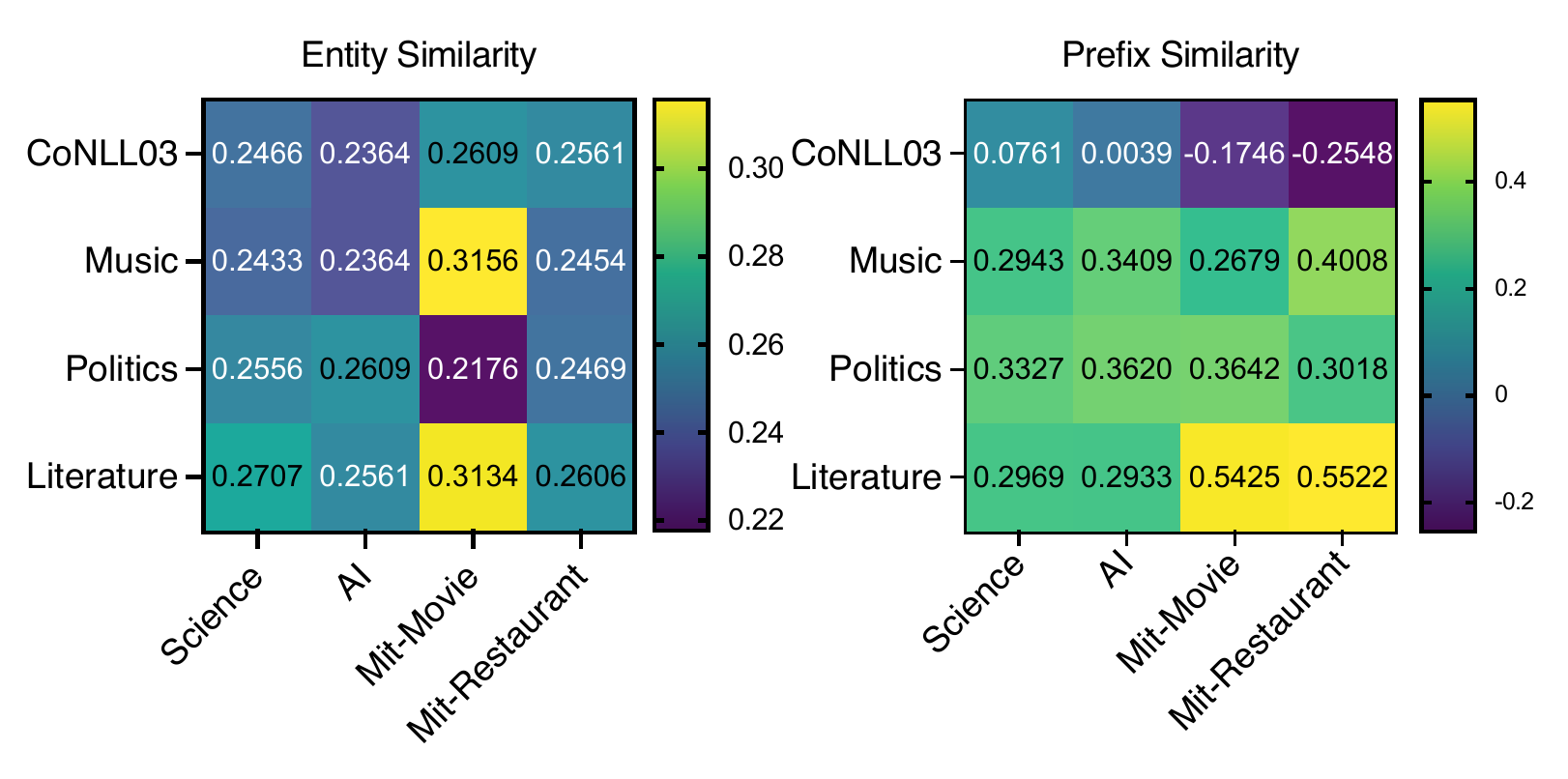}
    \caption{
    Similarity visualization of the source and target domains.
    }
    \label{fig:sim}
\end{figure}%

\subsection{Analysis}

\paragraph{Analysis of similarity-based selector.}
We visualize the entity and prefix similarity of the dual-query domain selector to analyze the correlation between source and target domains.
From Figure~\ref{fig:sim}, we find that the similarities in the left column are highly correlated with the entity semantic category. 
For example, the Music domain is most related to the target Movie domain, and they have the same entity category ``song’’ and similar entity categories ``genre'' and ``musicgenre’’. 
Furthermore, the entity similarity and the prefix similarity have similar trends, where the source domains with the highest similarity values are roughly the same. 
These observations reveal that the dual-query selector effectively identifies the most valuable source domains for the target domain.

\paragraph{Case analysis.} 
 As shown in Table~\ref{tab:case}, the models that are transferred from multiple source domains benefit from the source domains and perform more robustly. 
 For example, some entities have already appeared in the source domains and can be easily identified in the target domain. 
 Moreover, the same entities from source domains, such as ``Australia'' and ``Joseph'', provide direct support for the target domain, illustrating the effectiveness of our collaborative domain-prefix tuning.

\begin{figure}[!tpb] 
  \centering
    \includegraphics[width=0.48\textwidth]{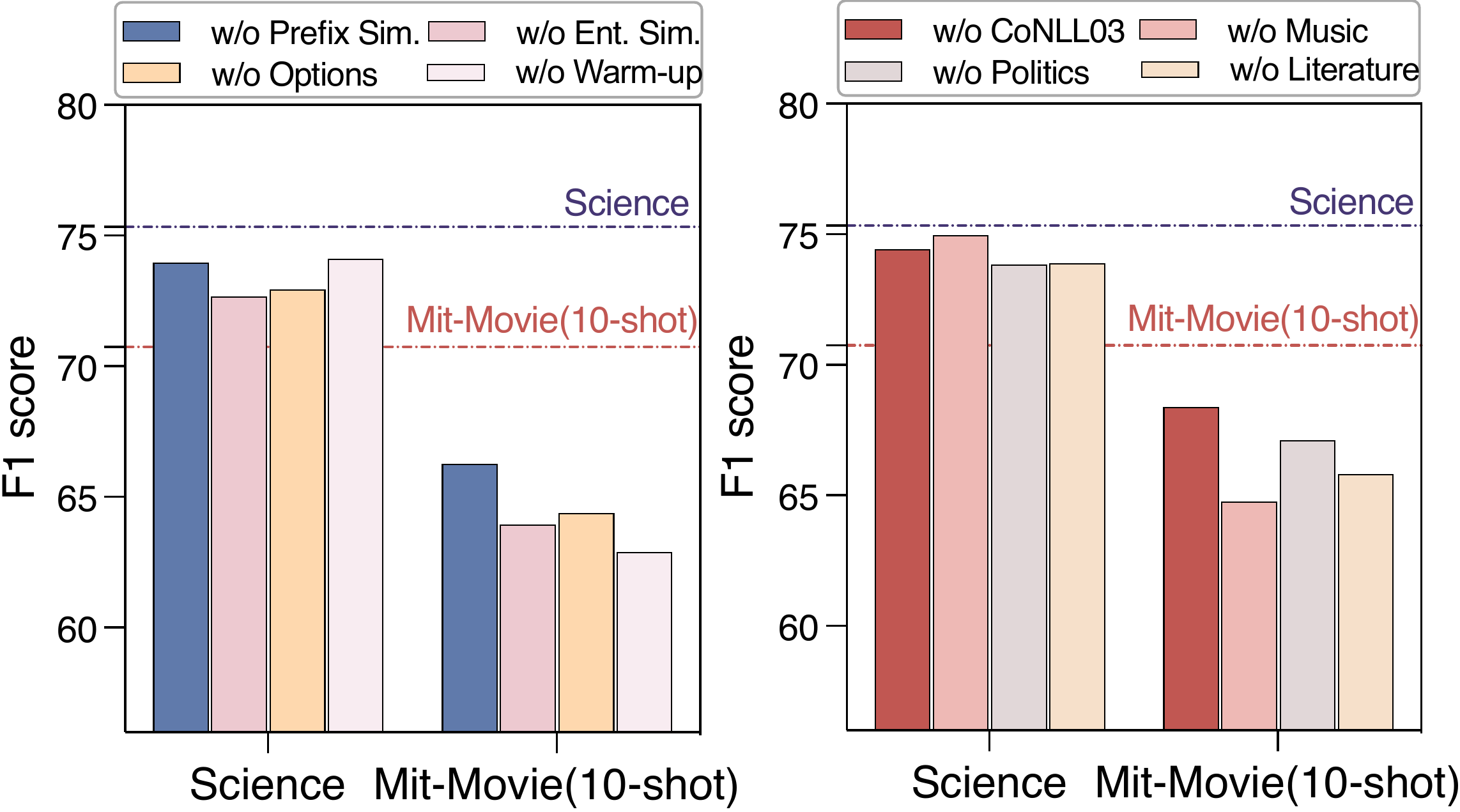}
    \caption{
        Ablation Study on the setting with multiple source domains. The dashed line denotes the results of our method in Table~\ref{tab:cross}.
    }
    \label{fig:ablation}
\end{figure}%

\paragraph{Ablation study.}  
As Figure.~\ref{fig:ablation} illustrates, we conduct an ablation study to evaluate the impact of various model components and source domains on domain transfer performance. 
Specifically, we systematically eliminate the entity similarity, prefix similarity, warm-up, and domain-related options from the model, respectively. 
Our results show that the performance of the models decreases significantly, particularly when the entity similarity component is removed. 
This highlights the crucial role played by our dual-query domain selector in facilitating effective domain transfer. 
Additionally, we investigate the effect of different source domains on the domain transfer performance. 
Our findings indicate that each source domain plays a unique role in the domain transfer process and exhibits varying levels of importance for different target domains, verifying the feasibility of leveraging knowledge from multiple domains to improve the target domain performance.

\section{Conclusion and Future Work}

In this paper, we target cross-domain NER and propose a lightweight approach dubbed {\ours}.
The proposed approach uses collaborative domain-prefix tuning to better utilize knowledge from multiple domains, thus, leading to better performance for on-the-fly cross-domain NER.
Through comprehensive experiments on benchmark datasets, we have shown that {\ours} achieves better results than a series of state-of-the-art cross-domain learning methods on NER.
Furthermore, from the perspective of the application, our {\ours} can realize one model for all domains (Keeping the base LM frozen), which is more in line with the needs of industrial applications.
In the future, we plan to extend the proposed approach to multilingual NER and other information extraction tasks. 
Besides, it is promising to build one unified large-scale generation model with pluggable and programmable modules to address the generalization issues.

\section*{Acknowledgments}
We would like to express gratitude to the anonymous reviewers for their kind comments. 
This work was supported by the National Natural Science Foundation of China (No.62206246 and U19B2027), Zhejiang Provincial Natural Science Foundation of China (No. LGG22F030011), Ningbo Natural Science Foundation (2021J190), and Yongjiang Talent Introduction Programme (2021A-156-G), CAAI-Huawei MindSpore Open Fund, and NUS-NCS Joint Laboratory (A-0008542-00-00). Our work was supported by Information Technology Center and State Key Lab of CAD\&CG, ZheJiang University.

\section*{Contribution Statement}
Xiang Chen, Lei Li and Shuofei Qiao make an equal contribution and share co-first authorship.
Ningyu Zhang and Huajun Chen are both the corresponding authors.

\bibliographystyle{named}
\bibliography{ijcai23}

\end{document}